\documentclass[journal]{IEEEtran}

\ifCLASSINFOpdf
\usepackage[pdftex]{graphicx}
\usepackage{subfigure}

\graphicspath{{../pdf/}{../jpeg/}{../png/}}

\DeclareGraphicsExtensions{.pdf,.jpeg,.png}
\else

\usepackage[dvips]{graphicx}

\graphicspath{{../eps/}}

\DeclareGraphicsExtensions{.eps}
\fi
\usepackage{amsmath}
\usepackage{array}

\usepackage{color} 
\usepackage{booktabs}
\usepackage{threeparttable}

\usepackage[top=2cm, bottom=2cm, left=2cm, right=2cm]{geometry}
\usepackage{algorithm}
\usepackage{algorithmicx}
\usepackage{algpseudocode}

\begin{document}
	
	\title{Communication-Efficient Federated Deep Learning with Asynchronous Model Update and Temporally Weighted Aggregation}

	\author{Yang Chen, 
		Xiaoyan Sun,
		Yaochu Jin,	
		\thanks{This work is supported by the National Natural Science Foundation of China with Grant No.61473298 and 61876184. (\textit{Corresponding author: Yaochu Jin})}
		\thanks{Y. Chen and Xi. Sun are with the School of Information and Control Engineering, China University of Mining and Technology, Xuzhou 221116, China. Y. Chen and X. Sun contributed equally to this work and are co-first authors.(e-mail: fedora.cy@gmail.com; xysun78@hotmail.com)}
		\thanks{Y. Jin is with the Department of Computer Science, University of Surrey, Guildford, GU2 7XH, United Kingdom. (Email: yaochu.jin@surrey.ac.uk)}}
	
	\markboth{}
	{Shell \MakeLowercase{\textit{et al.}}: Bare Demo of IEEEtran.cls for IEEE Journals}
	
	\maketitle
	
	\begin{abstract}
		Federated learning obtains a central model on the server by aggregating models trained locally on clients. As a result, federated learning does not require clients to upload their data to the server, thereby preserving the data privacy of the clients. One challenge in federated learning is to reduce the client-server communication since the end devices typically have very limited communication bandwidth. This paper presents an enhanced federated learning technique by proposing a synchronous learning strategy on the clients and a temporally weighted aggregation of the local models on the server. In the asynchronous learning strategy, different layers of the deep neural networks are categorized into shallow and deeps layers and the parameters of the deep layers are updated less frequently than those of the shallow layers. Furthermore, a temporally weighted aggregation strategy is introduced on the server to make use of the previously trained local models, thereby enhancing the accuracy and convergence of the central model. The proposed algorithm is empirically on two datasets with different deep neural networks. Our results demonstrate that the proposed asynchronous federated deep learning outperforms the baseline algorithm both in terms of communication cost and model accuracy.   
	\end{abstract}

	
	\IEEEpeerreviewmaketitle
	
	\section{INTRODUCTION}\label{sec1}
	
	Smart phones, wearable gadgets, and distributed wireless sensors usually generate huge volumes of privacy sensitive data. In many cases, service providers are interested in mining information from these data to provide personalized services, for example, to make more relevant recommendations to clients. However, the clients are usually not willing to allow the service provider to access the data for privacy reasons.\par
	
	Federated learning is a recently proposed privacy-preserving machine learning framework  \cite {mcmahan2017communication}. The main idea is to train local models on the clients, send the model parameters to the server, and then aggregate the local models on the server. Since all local models are trained upon data that are locally stored in clients, the data privacy can be perserved. The whole process of the typical federated learning is divided into communication rounds, in which the local models on the clients are trained on the local datasets. For the $k$-th client, where $k \in S$, and $S$ refers to the participating subset of $m$ clients, its training samples are denoted as $\mathcal{P}_k$ and the trained local model is represented by the model parameter vector $\omega^k$. In each communication round, only models of the clients belonging to the subset $S$ will download the parameters of the central model from the server ans use them as the initial values of the local models. Once the local training is completed, the participating clients send the updated parameters back to the server. Consequently, the central model can be updated by aggregating the updated local models, i.e. $\omega=Agg(\omega^k)$ \cite{Konecny2015,Konecny2016,mcmahan2017communication}.\par  
	
	In this setting, the local models of each client can be any type of machine learning models, which can be chosen according to the task to be accomplished. In most existing work on federated learning \cite{mcmahan2017communication}, deep neural networks (DNNs), e.g., long short-term memory (LSTM), are employed to conduct text-word/text-character prediction tasks.
	In recent years, DNNs have been successfully applied to many complex problem-solvings, including text classification, image classification, and speech recognition  \cite{lecun2015deep, Shin2016Deep, Greff2015LSTM}. Therefore, DNNs are widely adopted as the local model in federated learning, and the stochastic gradient descent (SGD) is the most popular learning algorithm for training the local models. \par 
	
	As aforementioned, one communication round includes parameter download (on clients), local training (on clients), trained parameter upload (on clients), and model aggregation (on the server). Such a framework appears to be similar to distributed machine learning algorithm \cite{ma2017distributed, reddi2016aide, shamir2014communication, zhang2015disco, chilimbi2014project, dean2012large}. In federated learning, however, only the models' parameters are uploaded and downloaded between the clients and server, and the data of local clients are not uploaded to the server or exchanged between the clients. Accordingly, the data privacy of each client can be preserved. \par 
	
	Compared with other machine leanring paradiagms, federated learning are subject to the following challenges  \cite{mcmahan2017communication, konevcny2016federated}:
	\begin{enumerate}
		\item \textbf{Unbalanced data}: The data amount on different clients may be highly imbalanced because there are light and heavy users. 
		\item \textbf{Non-IID data}: The data on the clients may be strongly non-IID because of different preferences of different users. As a result, local datasets are not able to represent the overall data distribution, and the local distributions are different from each other, too.  The IID assumption in distributed learning that training data are distributed over local clients uniformly at random \cite{Boyd2011a} usually does not hold in federated learning. 
		\item \textbf{Massively distributed data}: The number of clients is large. For example, the clients may be mobile phone users \cite{Konecny2015}, which can be enormous.  
		\item \textbf{Unreliable participating clients}: It is common that a large portion of participating clients are often offline or on unreliable connections. Again in case the clients are mobile phone users, their communication state can vary a lot and thus cannot ensure their participation in each round of learning \cite{mcmahan2017communication}.   
	\end{enumerate}
	
	Apart from the above challenges, the total communication cost is often used as an overall performance indicator of federated learning due to the limited bandwidth and battery capacity of mobile phones. Of course, like other learning algorithms, the learning accuracy,  which is mainly determined by the local training and the aggregation strategy, is also of great importance. Accordingly, the motivation of our work is to reduce the communication cost and improve the accuracy of the central model, assuming that DNNs are used as the local learning models. Inspired by interesting observations in fine-tuning of DNNs \cite{yosinski2014transferable}, an asynchronous strategy for local model updating and aggregation is proposed to improve the communication efficiency in each round.\par
	
	The main contributions of the present work are as follows. First, an asynchronous strategy that aggregates and updates the parameters in the shallow and deep layers of DNNs at different frequencies is proposed to reduce the number of parameters to be communicated between the server and clients. Second, a temporally weighted aggregation strategy is suggested to more efficiently integrate information of the previously trained local models in model aggregation to enhance the learning performance. \par  
	
	The remainder of the paper is organized as follows. In Section \ref{sec2}, related work is briefly reviewed. The detail of the proposed algorithm, especially the asynchronous strategy, the temporally weighted aggregation and the overall framework are described in Section \ref{sec3}. Section \ref{sec4} presents the experimental results and discussions. Finally, conclusions are drawn in Section \ref{sec5}.\par
	
	\section{Related Work}\label{sec2}
	Kone{\v{c}}n{\'{y}} et al. developed the first framework of federated learning and also experimentally proved that existing machine learning algorithms are not suited for this setting \cite{Konecny2015}. In \cite{Konecny2016}, Kone{\v{c}}n{\'{y}} et al. proposed two ways to reduce the uplink communication costs, i.e., structured updates and sketched updates, using data compression/reconstruction techniques. A more recent version of federated learning, FedAVG for short, was reported in \cite{mcmahan2017communication}, which was developed for obtaining a central prediction model of Google's Gboard APP and can be embedded in a mobile phone to protect the user's privacy. The pseudo code of FedAVG is provided in Algorithm 1. \par
	
	\begin{algorithm}
		\caption{FedAVG.}
		\label{AlgoFedAVG}
		\begin{algorithmic}[1]%do not show line#
			\Function {ServerExecution}{}  \Comment{Run on the server}
			\State initialize $w_{0}$
			\For{each round $t = 1,2,...$}
			\State $m \gets$ max($C\cdot K, 1$)
			\State $S_{t} \gets$ (random set of $m$ clients)
			\For{each client $k \in S_t$ \textbf{in parallel}}
			\State $w_{t+1}^k \gets$ ClientUpdate($k, w_t$)
			\EndFor
			\State $w_{t+1} \gets \sum_{k=1}^K \frac{n_k}{n} w_{t+1}^k$
			\EndFor
			\EndFunction
			\\
			\Function {ClientUpdate}{$k,w$} \Comment{Run on client $k$ }
			\State $\mathcal{B} \gets$ (split $\mathcal{P}_k$ into batches of size $B$)
			\For{each local epoch $i$ from 1 to $E$}
			\For{batch $b \in \mathcal{B}$}
			\State $w \gets$ $w- \eta \bigtriangledown \ell(w;b)$)
			\EndFor
			\EndFor
			\State return $w$ to server
			\EndFunction
		\end{algorithmic}
	\end{algorithm}
	
	In the following, we briefly explain the main components of FedAVG: 
	\begin{enumerate}
		\item \textbf{Server Execution} consists of the \textit{initialization} and \textit{communication rounds}. 
		\begin{enumerate}
			\item\textit{Initialization:} Line 2 initializes parameter $\omega_0$. \par
			\item\textit{Communication Rounds:} Line 4 obtains $m$, the number of participating clients; $K$ indicates the number of local clients, and $C$ corresponds to the fraction of participating clients per round, according to which line 5 randomly selects participating subset $S_t$. In lines 6-8, sub-function $Client Update$ is called in parallel to get $\omega_{t+1}^k$. 
			Line 9 executes aggregation to update $\omega_{t+1}$. \par
		\end{enumerate}
		\item \textbf{Client Update} The sub-function gets $k$ and $\omega$. $B$ and $E$ are the local mini-batch size and the number of local epochs, respectively. \par
		Line 14 splits data into batches. Lines 15-19 execute the local SGD on batches. Line 20 returns the local parameters.\par
	\end{enumerate}
	
	The equation in line 9, $\omega_{t+1} \gets \sum_{k=1}^K \frac{n_k}{n}*\omega_{t+1}^k$ described the aggregation strategy, where $n_k$ is the sample size of the $k$-th client and $n$ is the total number of training samples. $\omega_{t+1}^k$ comes from client $k$ in round $t+1$; however, it is not always updated in the current communication round. For clients that do not participate, their models remain unchanged until they are chosen to participate. In model aggregation, the parameters uploaded from clients in the current round and those in previous ones contribute equally to the central model. \par
	
	Apart from reducing communication costs, other studies of federated learning have focused on protocol security. To tackle differential attacks, for example, Gayer et al.  \cite{geyer2017differentially} proposed an algorithm incorporating a preserving mechanism into federated learning for client sided differential privacy. In \cite{bonawitz2017practical}, Bonawitz et al. designed an efficient and robust transfer protocol for secure aggregation of high-dimensional data.  \par
	
	While privacy preserving and reduction of communication costs are two important aspects to take into account in federated learning, the main concern remains to be the enhancement of learning performance of the central model on all data. The loss function of federated learning is defined as $ F\left( \omega  \right){\text{ = }}\sum\limits_{k = 1}^{K} {\frac{{{n_k}}}{n}{f_k}\left( \omega  \right)} $, where ${f_k}\left( \omega  \right)$ is the loss function of the $k$-th client model. Clearly, the performance of federated learning heavily depends on the model aggregation strategy. 
	
	Not much work has been reported on reducing the communication cost by reducing the number of parameters to be uploaded and downloaded between clients and the server except for some recent work reported most recently \cite{Zhu2018}. In this work, we present an asynchronous model learning mode that updates only part of the model parameters to reduce communication and a temporally weighted aggregation strategy that gives a higher weight on more recent models in aggregation to enhance learning performance.\par
	
	\section{Asynchronous Model Update and Temporally Weighted Aggregation}\label{sec3}
	\subsection{Asynchronous Model Update}\label{sec3A}
	The most intrinsic requirement for decreasing the communication cost of federated learning is to upload/download as little data as possible without deteriorating the performance of the central model. To this end, we present in this work an asynchronous model update strategy that updates only part of the local model parameters to reduce the amount of data to be uploaded or downloaded. 
	
	Our idea was primarily inspired from the following interesting observations made in fine tuning deep neural networks \cite{krizhevsky2014one,yosinski2014transferable}:
	\begin{enumerate}
		\item Shallow layers in a DNN learn the general features that are applicable to different tasks and datasets, meaning that a relatively small part of the parameters in DNNs (those in the shallow layers) represent features general to different data sets. 
		\item By contrast, deep layers in a deep neural network learn ad hoc features related to specific data sets and a large number of parameters focus on learning features in specific data. 
	\end{enumerate}
	
	These above observations indicate that the relatively smaller number of parameters of in the shallow layers are more pivotal for the performance of the central model in federated learning. Accordingly, parameters of the shallow layers should be updated more frequently than those parameters in the deep layers. Therefore, the parameters in the shallow and deep layers in the models can be updated asynchronously, thereby reducing the amount of data to be sent between the server and clients. We term this asynchronous model update. \par

\begin{figure}[!t]
		\centering
		\includegraphics[width=3.59in]{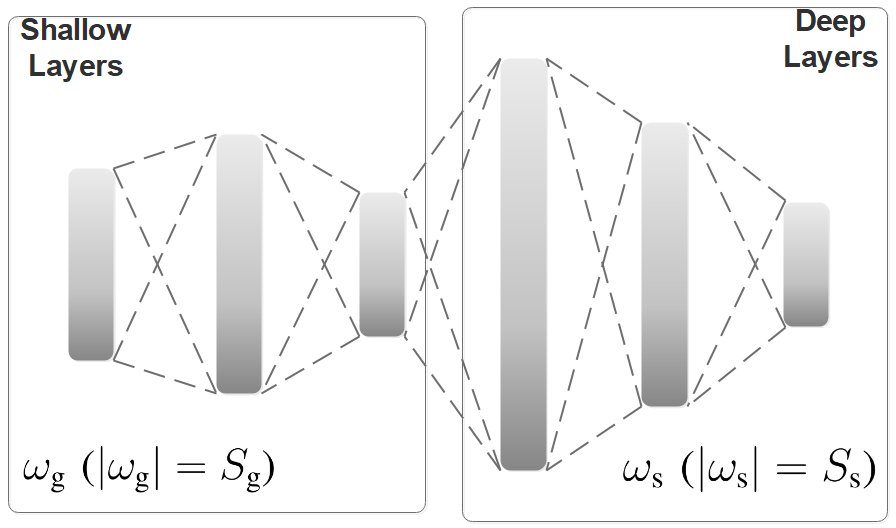}
		\caption{Illustration of shallow and deep layers of a deep neural network.}
		\label{fig_1}
	\end{figure}

	The DNN employed for the local models on the clients is shown in Fig. \ref{fig_1}. As illustrated in the figure, it can be separated into shallow layers for learning general features and deep layers for learning specific-feature layers features, which are denoted as  $\omega _\text{g}$ and $\omega _\text{s}$, respectively. The sizes of $\omega _\text{g}$ and $\omega _\text{s}$ are denoted as $S_\text{g}$ and $S_\text{s}$, respectively, and typically, $S_\text{g} \ll S_\text{s}$. In the proposed asynchronous learning strategy, $\omega _\text{g}$ will be updated and uploaded/downloaded more frequently than $\omega _\text{s}$. Assume that the whole federated process is divided into loops and each loop has $T$ rounds of model updates. In each loop, $\omega _\text{g}$ will be updated and communicated in every round, while $\omega _\text{s}$ will be updated and communicated in only $fe$ rounds, where $fe < T$. As a result, the number of parameters to be exchanged between the server and clients will be $(T-fe)*S_\text{s}$. This way, the communication cost can be significantly reduced, since $S_\text{s}$ is usually very large in DNNs. \par 
	
\begin{figure}[!t]
		\centering
		\subfigure[Synchronous model update strategy.]{
			\includegraphics[height=2.52in]{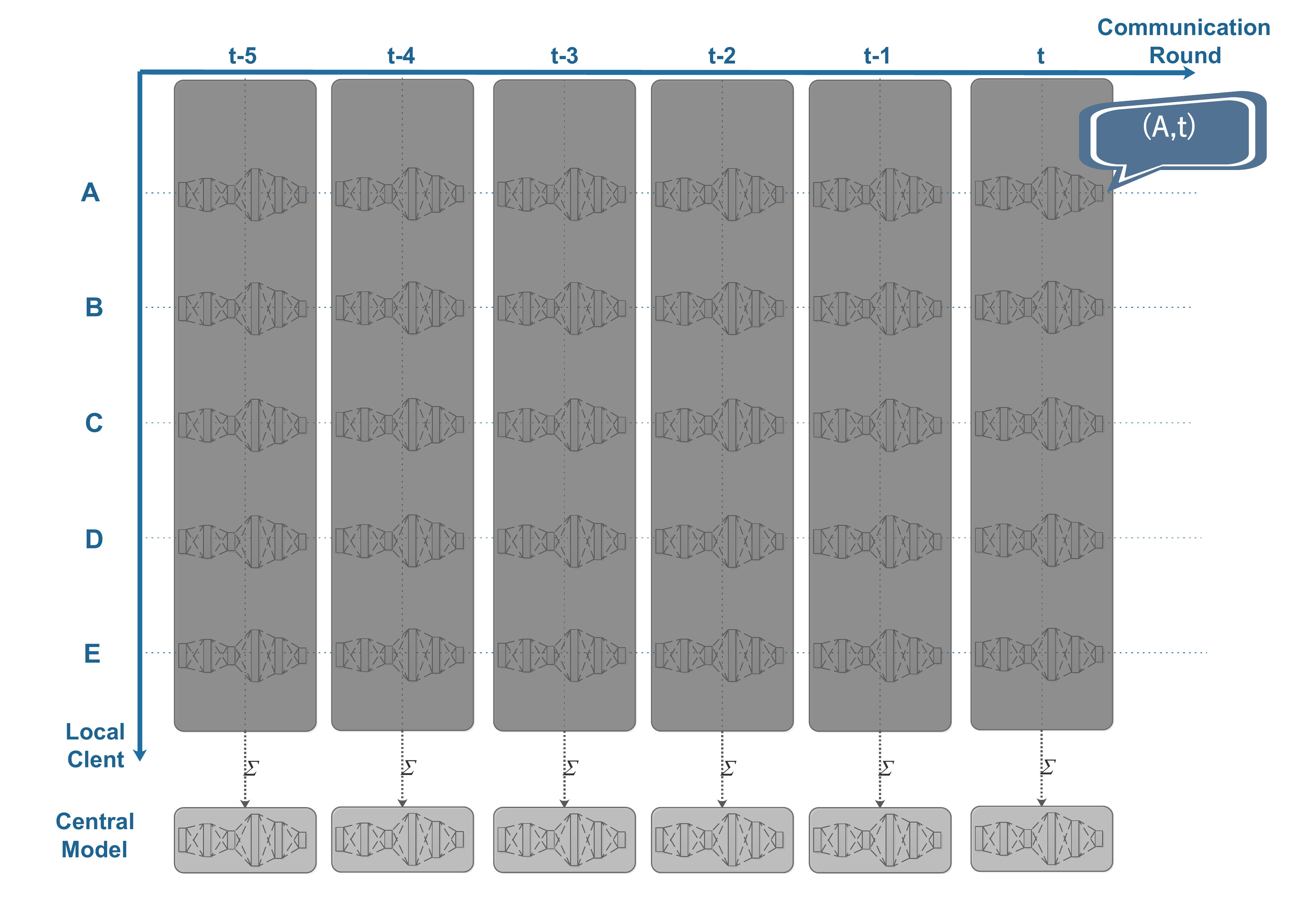}}
		\hspace{1in}
		\centering
		\subfigure[Asynchronous model update strategy.]{
			\includegraphics[height=2.52in]{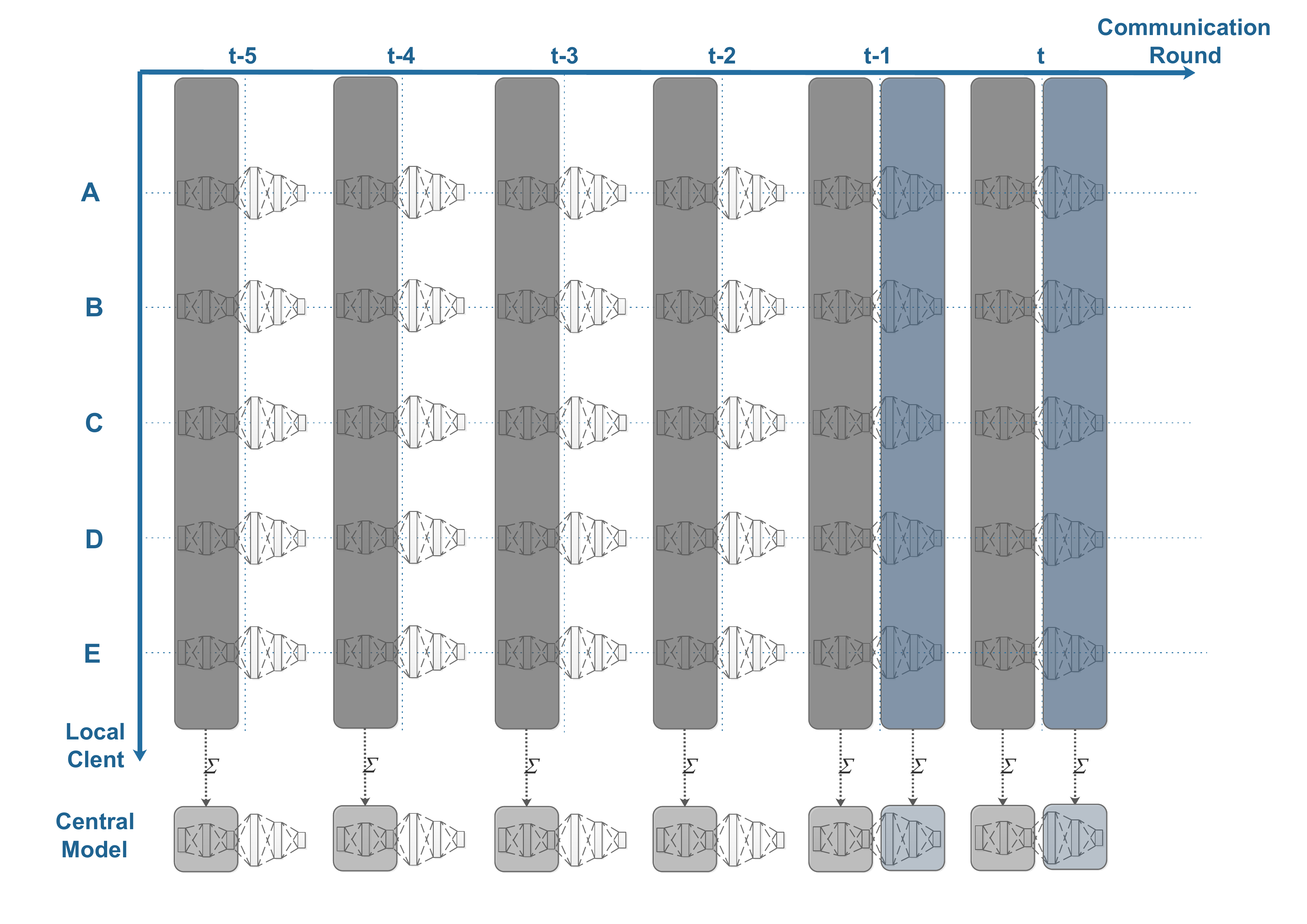}}
		\caption{Parameter Exchange.}
		\label{fig_2}
	\end{figure}

	An example is given in Fig. \ref{fig_2} to illustrate the asynchronous learning strategy. The abscissa and ordinate denote the communication round and the local client, respectively. In this example, there are five local devices, i.e., $\{A, B, C, D, E\}$ and a server. Point ($A, t$) indicates that client $A$ is participating in updating the global model in round $t$. 
	
	Fig. \ref{fig_2} (a) provides an illustration of a conventional synchronous aggregation strategy, where both $\omega _{{\text{g}}}$ and $\omega _{{\text{s}}}$ are uploaded/downloaded in each round. The grey rounded rectangles in the bottom represent the aggregation, in which both shallow and deep layers participate in all rounds. By contrast, Fig. \ref{fig_2} (b) illustrates the proposed asynchronous learning strategy. In this example, there are six computing rounds ($t-5, t-4, ..., t$) in the loop, and the parameters of deep layers are exchanged in rounds $t-1$ and $t$ only. As a result, the number of reduced parameters to be communicated is $ 2/3* S_{\text{s}}$.
	
	\subsection{Temporally Weighted Aggregation}\label{sec3B}
\begin{figure}[!t]
		\includegraphics[height=2.39in]{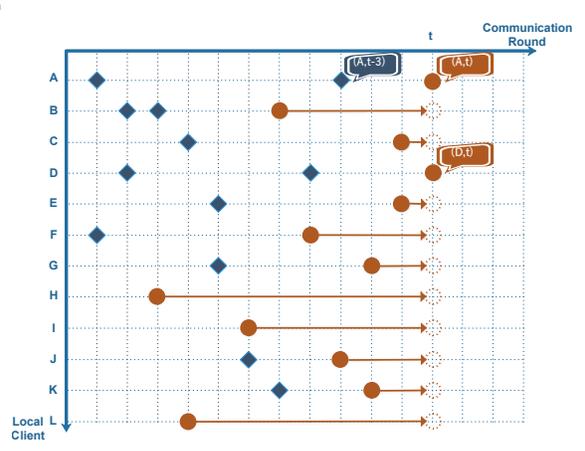}
		\caption{Conventional aggregation strategy.}
		\label{fig_3}
	\end{figure}

	In federated learning, the aggregation strategy (Line 9 in Algorithm \ref{AlgoFedAVG}) usually weights the importance of each local model based on the size of the training data on the corresponding client. That is, the larger $n_k$ is, the more the local on client $k$ will contribute to the central model, as illustrated in Fig. \ref{fig_3}. In that example, the blue diamonds represent the previously trained local models, which do not take part in the $t+1$-th update of the central model and only the orange dots are the most recently updated local model on each client and will participate in the current round of aggregation. All participating local models (orange dots in Fig. \ref{fig_3}) will be weighted by their data size only regardless the computing round in which these models are updated. In other words, local models updated in round $t-p$ are as important as those updated in round $t$, which might not be reasonable.
	
	In federated learning, however, the training data on each participating client are changing in each round and therefore, the most recently updated model should have a higher weight in the aggregation. Accordingly, the following model aggregation method taking into account of  timeliness of the local models is proposed. 
	\begin{figure}[!t]
		\includegraphics[height=2.36in]{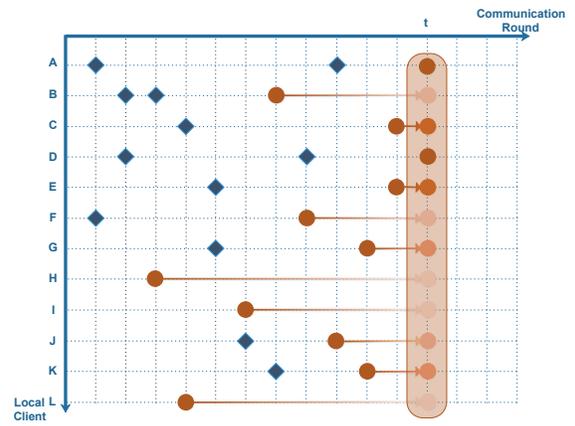}
		\caption{Illustration of the temporally weighted aggregation.}
		\label{fig_4}
	\end{figure}
	\begin{equation}\label{eq4}{
			\omega_{t+1} \gets \sum_{k=1}^K \frac{n_k}{n}*(e/2)^{-(t-timestamp^k)}*\omega^k
	}\end{equation}
	where $e$ is the natural logarithm used to depict the time effect, $t$ means the current round, and $timestamp^k$ is the round in which the newest $\omega^k$ was updated. The proposed temporally weighted aggregation is illustrated in Fig. \ref{fig_4}. Similar to the setting used in Fig. \ref{fig_3}, all clients participating in the aggregation are denoted by an orange dot, while others are represented by blue diamonds. Furthermore, the depth of the brown color indicates the timeliness and the deeper the color, the higher weight this local model will have in aggregation.
	
	\subsection{Framework}\label{sec3C}
	\begin{figure*}[!t]
		\centering
		\includegraphics[height=3.69in]{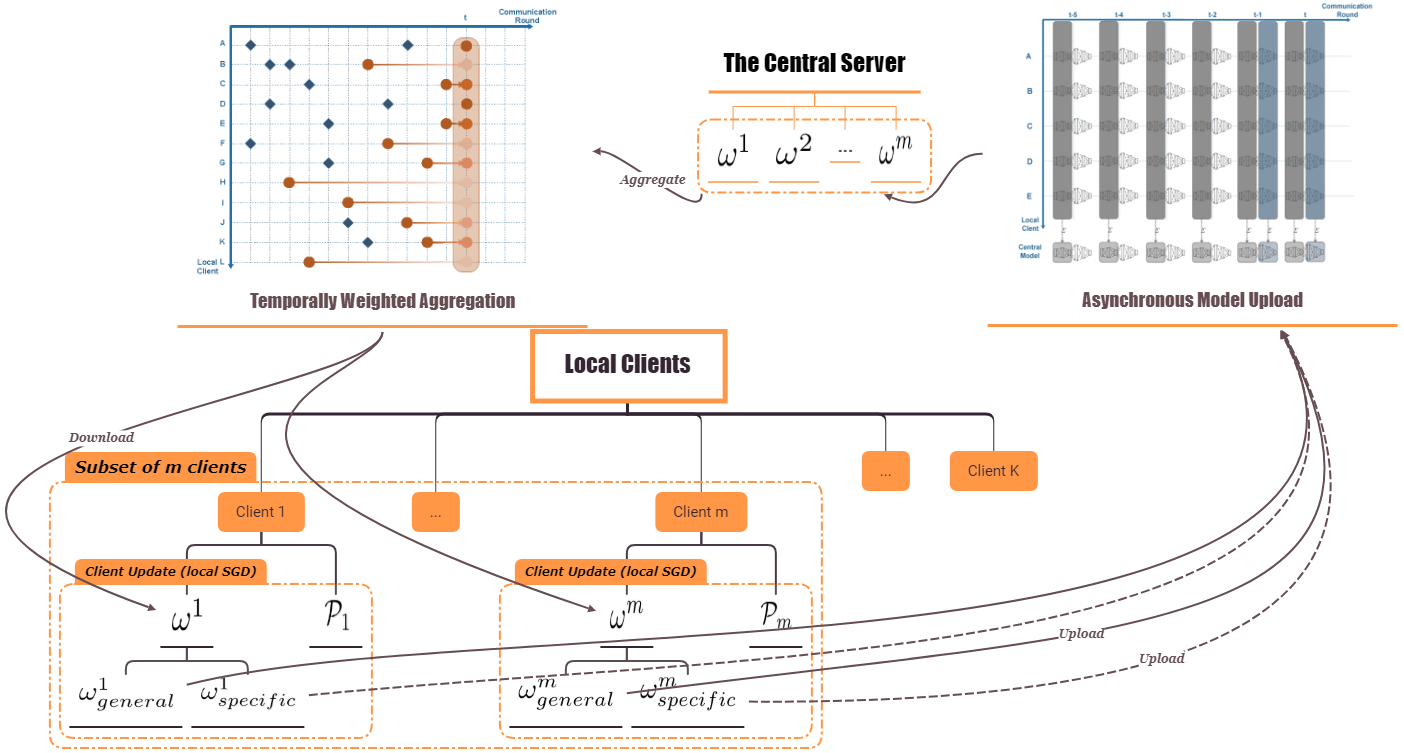}
		\caption{Federated learning with asynchronous model update and temporally weighted aggregation.}
		\label{fig_5}
	\end{figure*}
	The framework of the proposed federated learning with an asynchronous model update and temporally weighted aggregation is illustrated in Fig. \ref{fig_5}. A detailed description of the main components will be given in the following subsection.
	
	\subsection{Temporally Weighted Asynchronous Federated Learning}\label{sec3D}
	\begin{algorithm}[t]
		\caption{Server Component of TWAFL }
		\label{AlgoTWAFL_S}
		\begin{algorithmic}[1]%do not show line#
			\Function {ServerExecution}{}  \Comment{Run on the server}
			\State initialize $\omega_{0}$
			\For{each client $k \in \{1,2,...,K\}$ }
			\State $timestamp^k_\text{g} \gets 0$
			\State $timestamp^k_\text{s} \gets 0$
			\EndFor
			\For{each round $t = 1,2,...$}
			\If{$t$ mod $rounds\_in\_loop \in set_{ES}$}$^\S$ 
			\State $flag \gets$ True
			\Else
			\State $flag \gets$ False
			\EndIf
			\State $m \gets$ max($C * K, 1$)
			\State $S_{t} \gets$ (random set of $m$ clients)
			\For{each client $k \in S_t$ \textbf{in parallel}}
			\If{$flag$}
			\State $\omega^k \gets$ ClientUpdate($k, \omega_t, flag$)
			\State $timestamp^k_\text{g} \gets t$
			\State $timestamp^k_\text{s} \gets t$
			\Else
			\State $\omega^k_\text{g} \gets$ ClientUpdate($k, \omega_{g,t}, flag$)
			\State $timestamp^k_\text{g} \gets t$
			\EndIf
			\EndFor
			
			\State $\omega_{g,t+1} \gets \sum_{k=1}^K\frac{n_k}{n}*f_\text{g}(t,k)*\omega^k_\text{g}$ $^\dagger$
			\If{$flag$}
			
			\State $\omega_{s,t+1} \gets \sum_{k=1}^K\frac{n_k}{n}*f_\text{s}(t,k)*\omega^k_\text{s}$ $^\ddagger$
			\EndIf
			\EndFor
			\EndFunction
		\end{algorithmic}
		\begin{tablenotes}
			\item[]$\S$ $rounds\_in\_loop=15$ and $set_{ES}=\{11, 12, 13, 14, 0\}$
			\item[]$\dagger$ $f_\text{g}(t,k)=a^{-(t-timestamp^k_\text{g})}$
			\item[]$\ddagger$ $f_\text{s}(t,k)=a^{-(t-timestamp^k_\text{s})}$
		\end{tablenotes}
	\end{algorithm}
	
	The pseudo code for the two main components of the proposed temporally weighted aggregation asynchronous federated learning (TWAFL in short), one implemented on the server and the other on the clients is given in Algorithm \ref{AlgoTWAFL_S} and Algorithm \ref{AlgoTWAFL_C}, respectively. 
	
	\begin{algorithm}[!t]
		\caption{Client Component of TWAFL}
		\label{AlgoTWAFL_C}
		\begin{algorithmic}[1]%do not show line#
			\Function {ClientUpdate}{$k, w, flag$} \Comment{Run on client $k$ }
			\State $\mathcal{B} \gets$ (split $\mathcal{P}_k$ into batches of size $B$)
			\If{$flag$}
			\State $\omega \gets \omega$
			\Else
			\State $\omega_\text{s} \gets \omega$
			\EndIf
			
			\For{each local epoch $i$ from 1 to $E$}
			\For{batch $b \in \mathcal{B}$}
			\State $\omega \gets$ $\omega - \eta * \bigtriangledown \ell(w;b)$
			\EndFor
			\EndFor
			\If{$flag$}
			\State return $\omega$ to server
			\Else
			\State return $\omega_\text{s}$ to server
			\EndIf
			\EndFunction
		\end{algorithmic}
	\end{algorithm}
	
	The part to be implemented on the server consists of an initialization step followed by a number of communication rounds. In initialization (Algorithm \ref{AlgoTWAFL_S}, Lines 2-6 ),  the central model $\omega_0$, timestamps  $timestamp_\text{g}$ and $timestamp_\text{s}$ are initialized. Timestamps are stored and to be used to weight the timeliness of corresponding parameters in aggregation. \par
	
	The training process is divided into loops.	Lines 8-12 in Algorithm \ref{AlgoTWAFL_S} set $flag$ to be true in the last $1/freq$ rounds  in each loop. Assume there are $rounds\_in\_loop$ rounds in each loop. Lines 13-14 randomly select a participating subset $S_t$ of the clients according to $C$, which is the fraction of participating clients per round. In lines 15-24, sub-function $Client Update$ is called in parallel to get $\omega^k$/$\omega_\text{g}^k$, and the corresponding timestamps are updated. $flag$specified whether all layers or the shallow layers only will be updated and communicated. Then in lines 25-28, the aggregation is performed to update $\omega_\text{g}$. Note that compared with Equation \ref{eq4}, a parameter $a$ is introduced into the weighting function in line 25 or line 27 to examine the influence of different weightings in the experiments. In this work, $a$ is set to $e$ or $e/2$.  \par
	
	The implementation of the local model update (Algorithm \ref{AlgoTWAFL_C}) is controlled by three parameters, $k$, $\omega$, and $flag$, where $k$ is the index of the selected client, $flag$ indicates whether all layers or the shallow layers will be updated. $B$ and $E$ denote the local mini-batch size and the local epoch, respectively. In Algorithm \ref{AlgoTWAFL_C}, Line 2 splits data into batches, whereas Lines 3-7 set all layers or shallow layers of the local model to be downloaded according to $flag$. In lines 8-12, local SGD is performed. Lines 13-17 return local parameters.\par
	
	\section{Experimental Results and Analyses}\label{sec4}
	
	\subsection{Experimental Design}\label{sec4A}
	We perform a set of experiments using two popular deep neural network models, i.e., convolution neural networks (CNNs) for image processing and long short-term memory (LSTM) for human activity recognition, respectively, to empirically examine the learning peformance and communication cost of the prposed TWAFL. A CNN with two stacked convolution layers and one fully connected layer \cite{krizhevsky2012imagenet,lecun1998gradient} is applied on the MNIST handwritten digit recognition data set, and a LSTM with two stacked LSTM layers and one fully connected layer is chosen to accomplish the human activity recognition task \cite{anguita2013public,gers1999learning}. Both MNIST and the human activity recognition datasets are adapted to test the performance of the proposed federated learning framework for different real-world scenarios, e.g., non-IID distribution, unbalanced amount, and massively decentralized datasets, which will be discussed in detail in the next subsection. 
	
	The federated averaging (FedAVG) \cite{mcmahan2017communication} is selected as the baseline algorithm since it is the state-of-the-art approach. The proposed TWAFL is also compared with two variants, namely, TWFL that adopts temporally weighted aggregation without asynchronous model update, and AFL that employs the asynchronous model update without using temporally weighted aggregation. Thus, four algorithms, i.e., FedAVG, TWAFL, TEFL and AFL will be compared in the following experiments. \par
	
	The most important parameters in the proposed algorithm are listed in Table \ref{Table1}. Parameter $freq$ controls the frequency for updating and exchanging the parameters in the deep layers $\omega_\text{s}$ between the server and the local clients in a loop. For instance, $freq = 5/15$ means that only in the last five of the 15 rounds, the parameters in the deep layers $\omega_\text{s}$ will be uploaded/downloaded between the server and clients. $a$ is a parameter for adjusting the time effect in model aggregation. $K$ and $m$ are environmental parameters controlling the scale or complexity of the experiments, $K$ denotes the number of local clients, and $m$ is the number of participating clients per round. \par 
	
	\begin{table}[!t]
		\centering
		\caption{Parameter settings}
		\label{Table1}
		\begin{threeparttable}
			\begin{tabular}{@{}p{3.6cm}<{\centering}p{3.6cm}<{\centering}@{}}
				\toprule
				\textbf{Notion} & \textbf{Parameter Range} \\ \midrule
				$freq$ & \{3/15, $5/15^*$, 7/15\} \\ 
				$a$ & \{$e/2^*$,  e\} \\ 
				$K$ & \{10, $20^*$\} \\
				$m$ & \{1, $2^*$\} \\
				\bottomrule
			\end{tabular}
			\begin{tablenotes}
				\item[*] Default setting
			\end{tablenotes}
		\end{threeparttable}
	\end{table}
	
	\subsection{Settings on Datasets}\label{sec4B}
	As discussed in Section I, the federated learning framework has its particular challenges, such as non-IID, unbalanced, and massively decentralized datasets. Therefore, the datasets used in our experiments should be designed to reflect these challenges. The generation of the client dataset is described in detail in Algorithm \ref{AlgoG_localData}, which is controlled by four parameters $Labels$, $N_c$, $S_{min}$, and $S_{max}$, where $N_c$ controls the number of classes in each local dataset, $S_{min}$ and $S_{max}$ specify the range of the size of the local data, and $Labels$ indicates the names of classes involved in the corresponding tasks. 
	
	\begin{algorithm}
		\caption{Generation of Local Datasets.}
		\label{AlgoG_localData}
		\begin{algorithmic}[1]%do not show line#
			\Require{$Labels$, $N_C$, $S_{min}$, and $S_{max}$}
			\Ensure{non-IID and unbalanced local dataset $\mathcal{P}_k$}
			\State $classes \gets$ Choices($Labels$,$N_C$)
			\State $L \gets $ Len($Labels$)
			\State $weights \gets$ Zeros($L$)
			\State $\mathcal{P} \gets$ Zeros($L$)
			\For{each $class \in classes$}
			\State	$weights_{class} \gets$ Random(0,1)
			\EndFor
			\State $sum \gets  \sum_{class=1}^{L}  weights_{class}$
			\State $num \gets $ Random($S_{min},S_{max}$) 
			\For{each $class \in classes$}
			\State	$\mathcal{P}_{class} \gets \frac{weights_{class}}{sum} \times num $ 
			\EndFor
			\State $\mathcal{P}_{k} \gets \mathcal{P}$
		\end{algorithmic}
	\end{algorithm}
	
	\subsubsection {Handwritten Digit Recognition Using CNN}\label{sec4B1}
	
\begin{figure}[h]
		\centering
		\subfigure[1@MNIST.]{
			\includegraphics[height=2.36in]{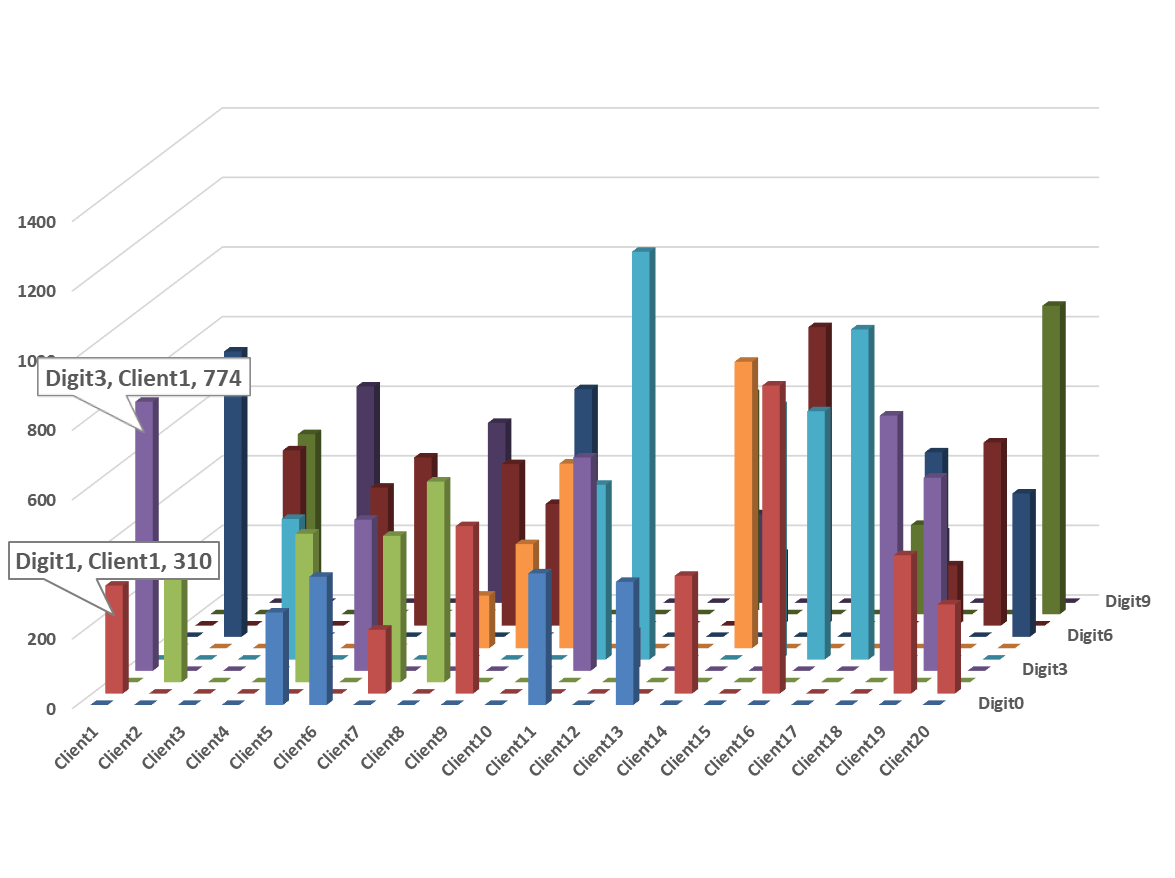}}
		\subfigure[2@MNIST.]{
			\includegraphics[height=1.16in]{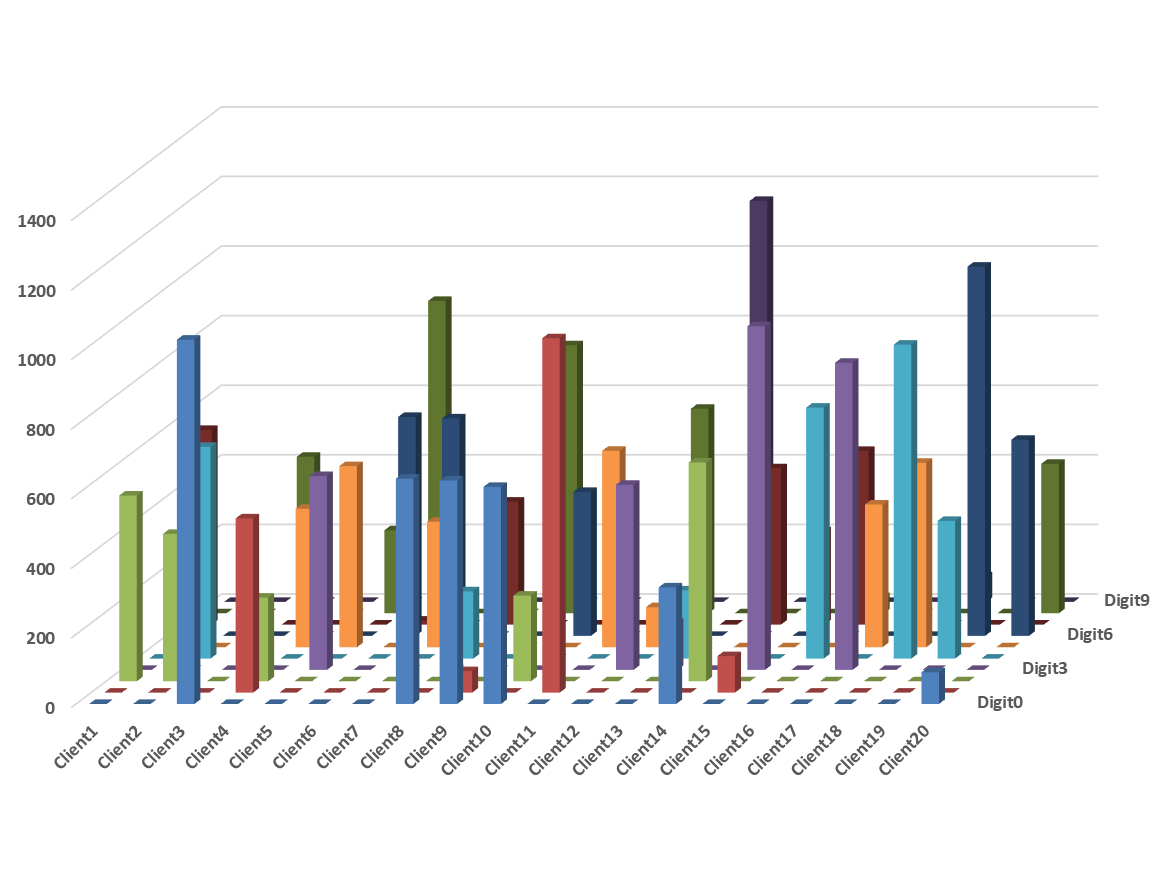}}
		\subfigure[3@MNIST.]{
			\includegraphics[height=1.16in]{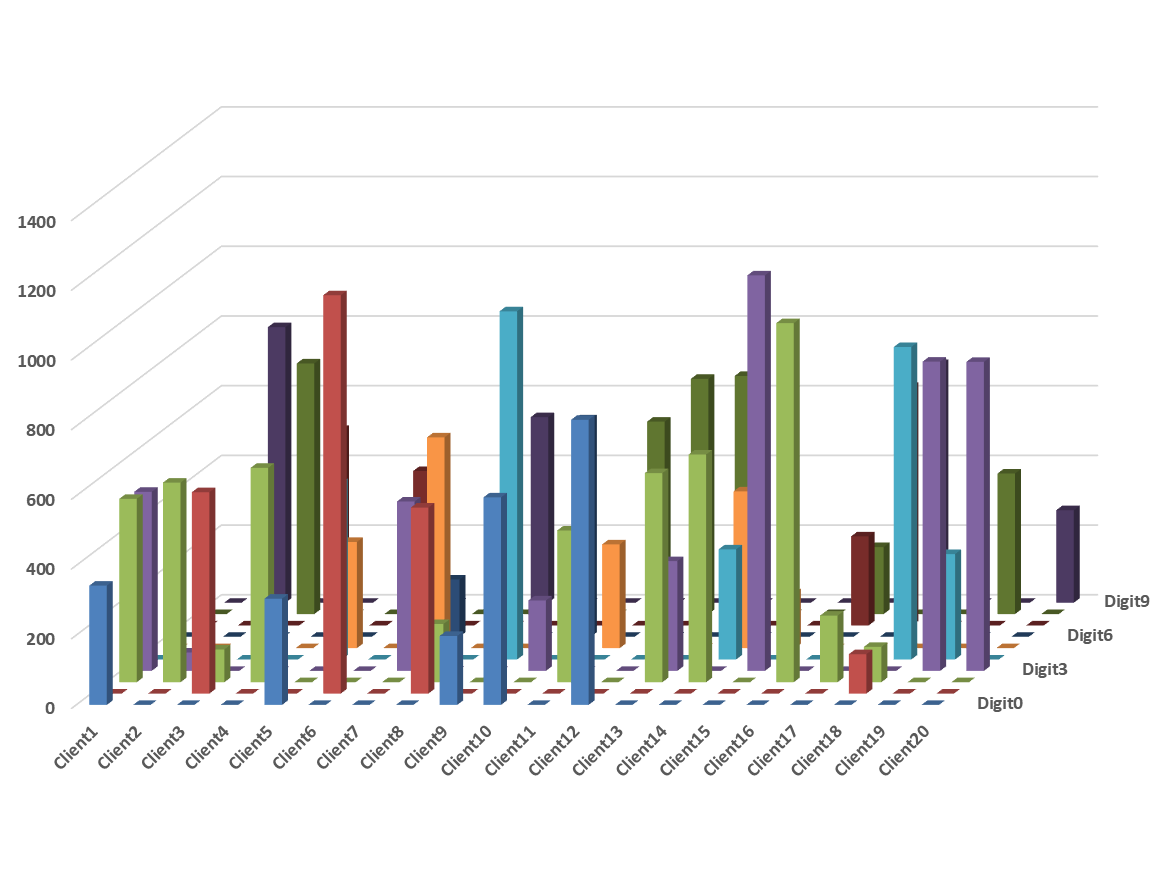}}
		\subfigure[4@MNIST.]{
			\includegraphics[height=1.16in]{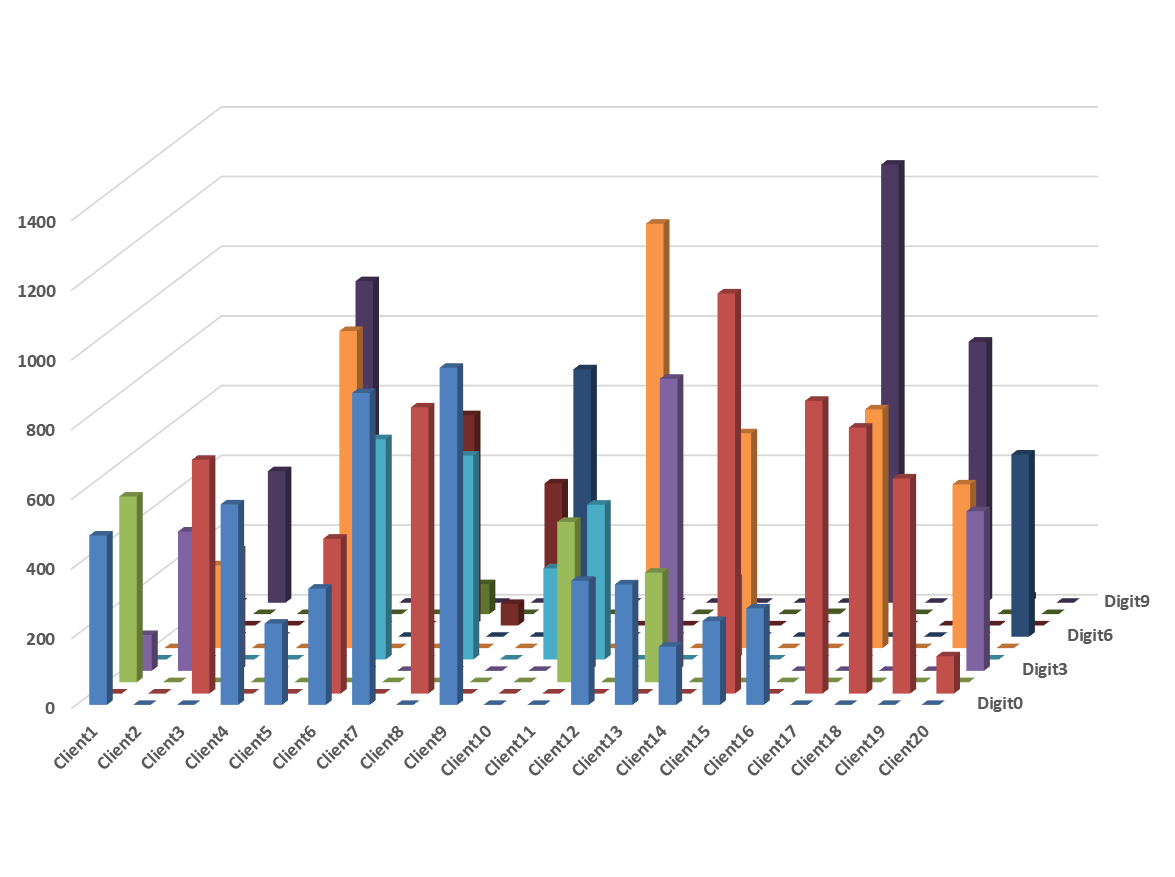}}
		\subfigure[5@MNIST.]{
			\includegraphics[height=1.16in]{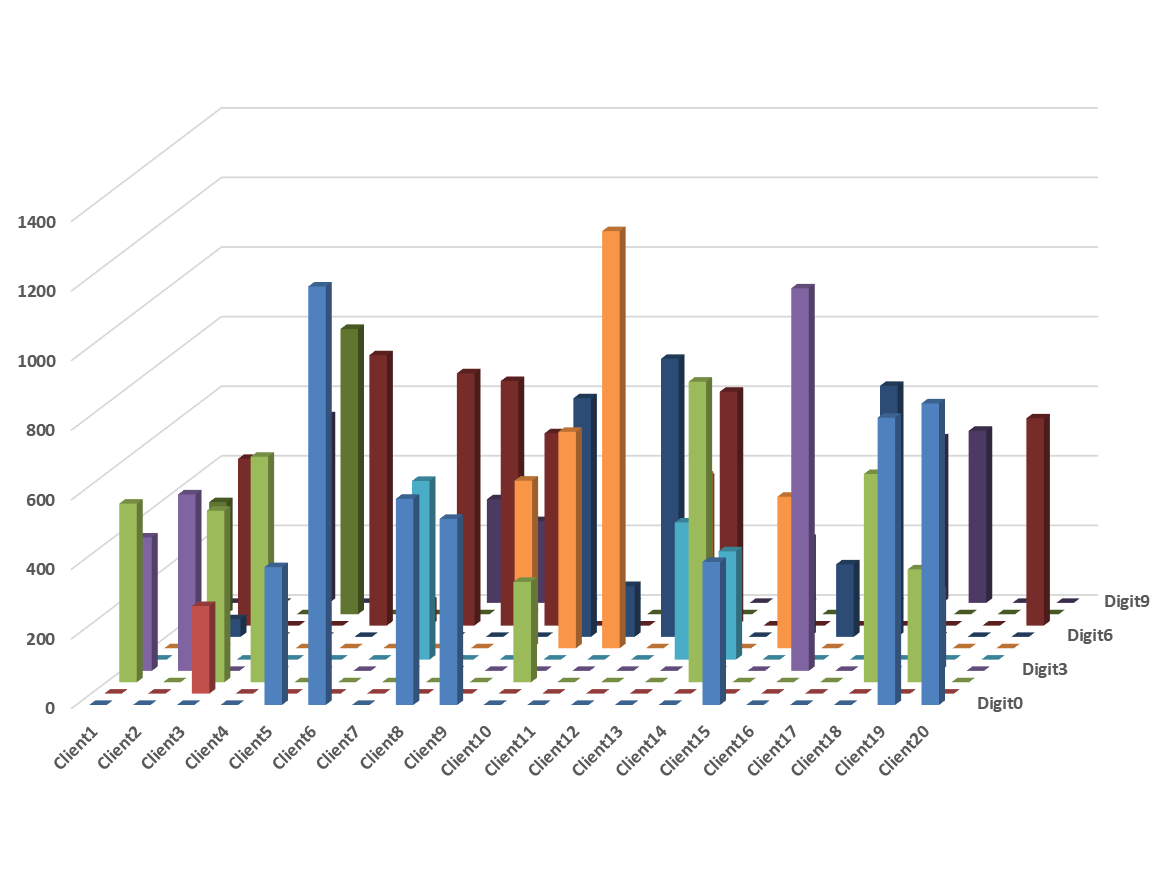}}
		\caption{3-D Column Charts of Pre-Generated Datasets.}
		\label{fig_6abcde}
	\end{figure}

	The MNIST dataset has ten different kinds of digits and one digit is a 28*28-pixel gray-scale image. To partition the data over local clients, we first sort them by their digit labels and divide them into ten shards, i.e., 0, 1, ..., 9. Then, Algorithm \ref{AlgoG_localData} is performed to compute $\mathcal{P}_k$, which is the $k$-th client's partition coefficient corresponding to these shards. In this task, $Labels = \{0, 1, 2, ..., 9\}$; $N_c$ is randomly chosen from $\{2, 3\}$, given $K = 20$, $S_{min} = 1000$ and $S_{max} = 1600$. For the sake of easy analyses, five partitions/local datasets, namely 1@MNIST, 2@MNIST, ..., 5@MNIST are predefined. Their corresponding 3-D column charts are plotted in Fig. \ref{fig_6abcde}. 
	
	The architecture of the CNN used for the MNIST task has two $5\times5$ convolution layers (the first one has 32 channels and the other has 64) and $2\times2$ max pooling layers. Then a fully connected layer with 512 units and ReLu activation is followed. The output layer is a softmax unit. The parameter settings for the CNN are listed in Table \ref{Table2}. \par
	
	\begin{table}[!t]
		\centering
		
		\caption{Parameters settings for the CNN.}
		\label{Table2}
		\begin{tabular}{@{}p{3.6cm}<{\centering}p{3.6cm}<{\centering}@{}}
			\toprule
			\textbf{Layer} & \textbf{Shapes} \\ \midrule
			conv2d\_1 & $5\times5\times1\times32$ \\
			conv2d\_1 & $32$ \\
			conv2d\_2 & $5\times5\times32\times64$ \\
			conv2d\_2 & $64$  \\
			dense\_1 & $1024\times512$ \\ 
			dense\_1 & $512$ \\
			dense\_2 & $512\times10$ \\ 
			dense\_2& $10$ \\ \bottomrule
		\end{tabular}
	\end{table}
	
	\subsubsection{Human Activity Recognition using LSTM}\label{sec4B2}
	In the Human Activity Recognition (HAR) dataset, each data is a sequence of images with a label out of six activities. Similar operations are applied on the HAR dataset to divide the dataset over local clients. Here, $Labels = \{0, 1, 2, ..., 5\}$; $N_c$ is randomly chosen from $\{2, 3\}$, given $K = 20$, $S_{min} = 250$ and $S_{max} = 500$. \par
	
	The architecture of the LSTM used in this study has two $5\times5$ LSTM layers (the first one with $cell\_size = 20$ and $time\_steps = 128$ and the other with $cell\_size = 10$ and the same $time\_steps$), a fully connected layer with 128 units and the ReLu activation, and a softmax output layer. The corresponding parameters of the LSTM is given in Table  \ref{Table3}.  \par
	
	\begin{table}[!t]
		\centering
		
		\caption{Parameter settings for the LSTM.}
		\label{Table3}
		\begin{tabular}{@{}p{3.6cm}<{\centering}p{3.6cm}<{\centering}@{}}
			\toprule
			\textbf{Layer} & \textbf{Shapes} \\ \midrule
			lstm\_1 & $9\times100$ \\
			lstm\_1 & $25\times100$ \\
			lstm\_1 & $100$ \\
			lstm\_2 & $25\times100$  \\
			lstm\_2 & $25\times100$ \\ 
			lstm\_2 & $100$ \\
			dense\_1 & $25\times256$ \\ 
			dense\_1& $256$ \\ 
			dense\_2 & $256\times6$ \\ 
			dense\_2& $6$ \\ \bottomrule
		\end{tabular}
	\end{table}
	
	\subsection{Results and analysis}\label{sec4C}
	Two sets of experiments are performed in this subsection. The first set of experiments examines the influence of the most important parameters, $freq$, $a$, $K$ and $m$, on the performance of the two strategies using the CNN for the 1@MNIST dataset. The second set of experiments compares four algorithms in terms of the communication cost and learning accuracy using the LSTM on the HAR dataset, since the HAR is believed to be more challenging than the MNIST dataset. 
	
	\subsubsection{Effect of the Parameters}\label{sec4c1}
	The experiments here are not meant for a detailed sensitivity analysis. These experiments and related discussions aim to offer a basic understanding of parameter settings that may be helpful in practice. We mainly conduct research on the parameters listed in Table \ref{Table1}. \par 
	
	In investigating the influence of a particular parameter, all others are set to be their default value. Experiment on $freq$ are carried out for  $freq=\{3/15, 5/15, 7/15\}$, given $a = e/2$, $K = 20$, and $m = 2$. \par
	
	Two metrics are adopted in this work for measuring the performance of the compared algorithms. One is the best accuracy of the central model within 200 rounds and the other is to the required rounds before the central model's accuracy reaches 95.0\%. Note that the same computing rounds means the same communication cost. All experiments are independently run for ten times, and their average (AVG) and standard deviation (STDEV) values are presented in Tables \ref{Table5}, \ref{Table6}, and \ref{Table8}. In the tables, the average value is listed before the standard deviation in parenthesis. \par
	
	\begin{table}[!t]
		\centering
		\caption{Experimental results on $freq$.}
		\label{Table5}
		\begin{threeparttable}
			\begin{tabular}{@{}p{2.4cm}<{\centering}p{2.4cm}<{\centering}p{2.4cm}<{\centering}@{}}
				\toprule
				\textit{\textbf{freq}} & \textbf{Accuracy\tnote{*}} & \textbf{Round\tnote{**}} \\ \midrule
				3/15 & 96.72\% (0.0037) & 115.74 (32.19)  \\
				5/15 & 97.08\% (0.0037) & 87.82 (20.03) \\
				7/15 & 97.12\% (0.0046) & 95.43 (14.64)  \\
				\bottomrule
			\end{tabular}
			\begin{tablenotes}
				\item[*] AVG (STDEV) of best overall accuracy within 200 rounds.
				\item[**] TWAFL on 1@MNIST reaches the accuracy 95.0\% within 200 rounds.
			\end{tablenotes}
		\end{threeparttable}
	\end{table}
	
	\begin{table}[!t]
		\centering
		\caption{Experimental results on $a$.}
		\label{Table6}
		\begin{threeparttable}
			\begin{tabular}{@{}p{2.4cm}<{\centering}p{2.4cm}<{\centering}p{2.4cm}<{\centering}@{}}
				\toprule
				\textit{\textbf{a}} & \textbf{Accuracy\tnote{*}} & \textbf{Round\tnote{**}} \\ \midrule
				e\tnote{***} &   96.92\% (0.0041) & 75.26 (2.09) \\
				e/2 &   97.08\% (0.0037) & 87.82 (20.03) \\
				\bottomrule
			\end{tabular}
			\begin{tablenotes}
				\item[*] AVG (STDEV) of best overall accuracy within 200 rounds.
				\item[**] TWAFL on 1@MNIST reaches the accuracy 95.0\% within 200 rounds.
				\item[***] e $\approx$ 2.72
			\end{tablenotes}
		\end{threeparttable}
	\end{table}
	
	Based on the results in Tables \ref{Table5}, \ref{Table6}, and \ref{Table8}, the following observations can be made.
	\begin{itemize}
		\item \textbf{Analysis on \textit{freq}:} From the results presented in Table \ref{Table5}, we can conclude that the lower the exchange frequency is, the less communication costs will be required, which is very much expected. However, a too low $freq$ will deteriorate the accuracy of the central model. \par
		\item \textbf{Analysis on \textit{a}:} $a$ is a parameter that controls the influence of time effect on the model aggregation. When $a$ is set as $e$, the more recently updated local models are more heavily weighted in the aggregated model, and $a$ takes the value $e/2$, the previously updated local models will have a greater impact on the central model. The results in Table \ref{Table6} indicate that $e/2$ is a better option for the CNN on the 1@MNIST dataset.  When $a=1$, the algorithm is reduced to AFL, meaning that the parameters uploaded in different rounds will be of equal importance in the aggregation.  \par
	\end{itemize}	
	
	Both parameters $K$ and $m$ leverage the scalability of federated learning. The AVG and STDEV values are calculated based on the recognition accuracy of the CNN on the five predefined datasets. Different combinations of $K$ and $m$ are separately assessed. Based on the results presented in Table \ref{Table8}, the following three conclusions can be made. 
	First, a larger number of involved clients (a larger $m$) leads to a higher recognition accuracy. Second, TWAFL outperforms FedAVG when the active client fraction is $C = 0.1$, as indicated in the results in rows one and two in the table. Third, FedAVG is slightly better when the total number of clients is smaller and $C$ is higher, as shown by the results listed in the last row of the table. This implies that the advantage of the proposed algorithm over the traditional federated learning will become more apparent as the number of  clients increases. This is encouraging since in most real-world problems, the number of clients is typically very large.	
	
	\begin{table}[!t]
		\centering
		\caption{Experimental results on the scalability of the algorithm.}
		\label{Table8}
		\begin{threeparttable}
			\begin{tabular}{@{}p{2.4cm}<{\centering}p{2.4cm}<{\centering}p{2.4cm}<{\centering}@{}}
				\toprule
				\begin{tabular}{@{}c@{}}
					\textbf{Scalability} \\
					\midrule
					\begin{tabular}{@{}cc@{}}
						\textit{\textbf{K}} & \textit{\textbf{m}} \\
					\end{tabular}
				\end{tabular}	&
				\textbf{FedAVG}  &
				\textbf{TWAFL} \\
				\midrule
				\begin{tabular}{@{}cc@{}}
					20 & 2 \\
				\end{tabular} &
				96.83\% (0.0097) &
				97.16\% (0.0096) \\
				\begin{tabular}{@{}cc@{}}
					10 & 1 \\
				\end{tabular}& 
				94.26\% (0.0083) &
				94.76\% (0.0102) \\
				\begin{tabular}{@{}cc@{}}
					10 & 2 \\
				\end{tabular} & 
				96.16\% (0.0167)  &
				96.11\% (0.0909) \\
				\bottomrule
			\end{tabular}
			\begin{tablenotes}
				\item[*] AVG (STDEV) of best overall accuracy within 200 rounds.
			\end{tablenotes}
		\end{threeparttable}
	\end{table}
	
	\subsubsection{Comparison on Accuracy and Communication Cost}\label{sec4c2}
	\begin{figure}[h]
		\centering
		\subfigure[1@MNIST.]{
			\includegraphics[height=1.86in]{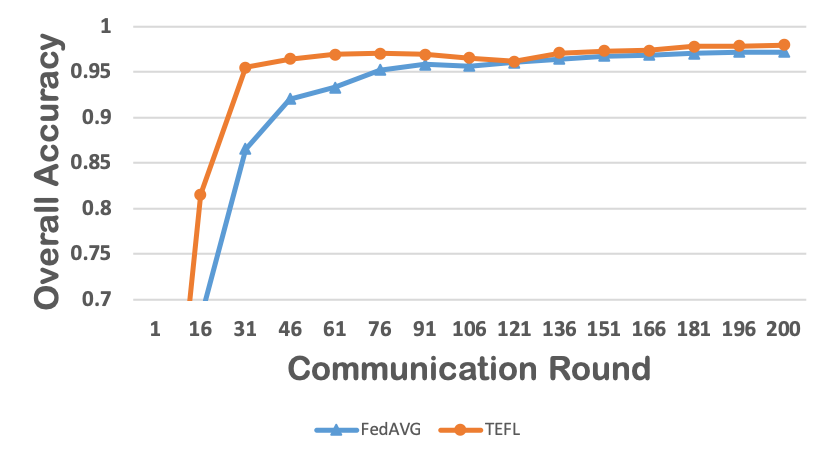}}
		\subfigure[2@MNIST.]{
			\includegraphics[height=0.86in]{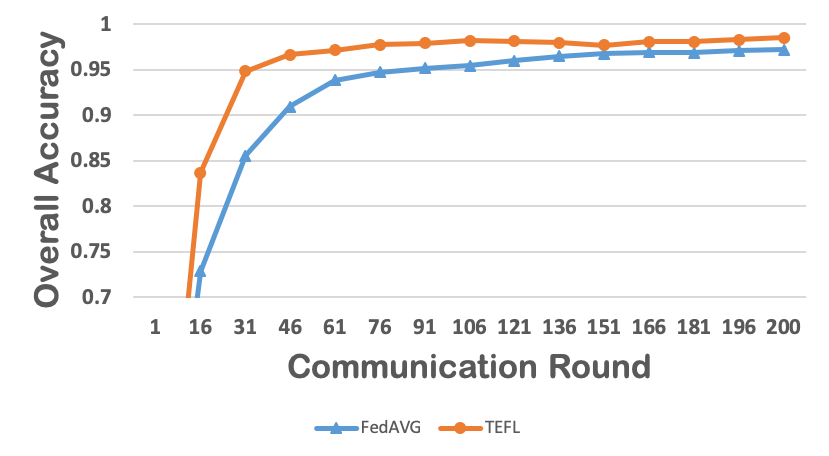}}
		\subfigure[3@MNIST.]{
			\includegraphics[height=0.86in]{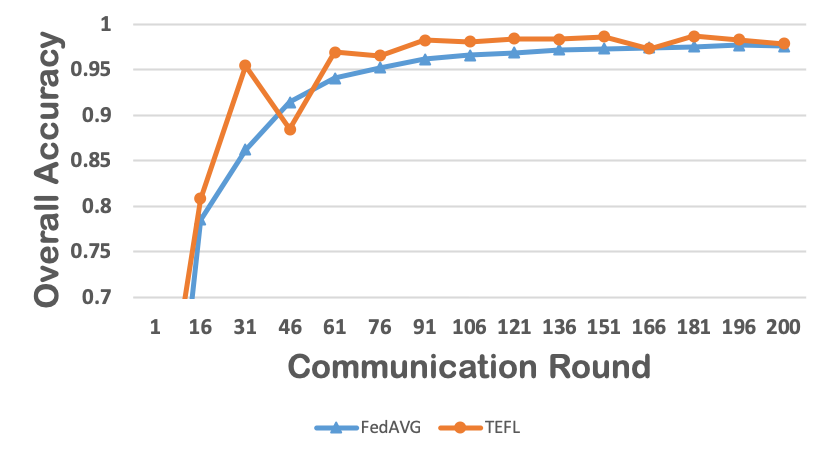}}
		\subfigure[4@MNIST.]{
			\includegraphics[height=0.86in]{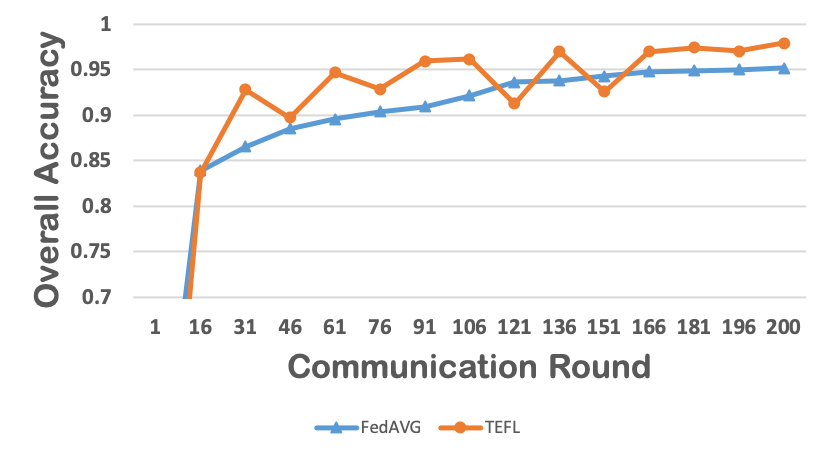}}
		\subfigure[5@MNIST.]{
			\includegraphics[height=0.86in]{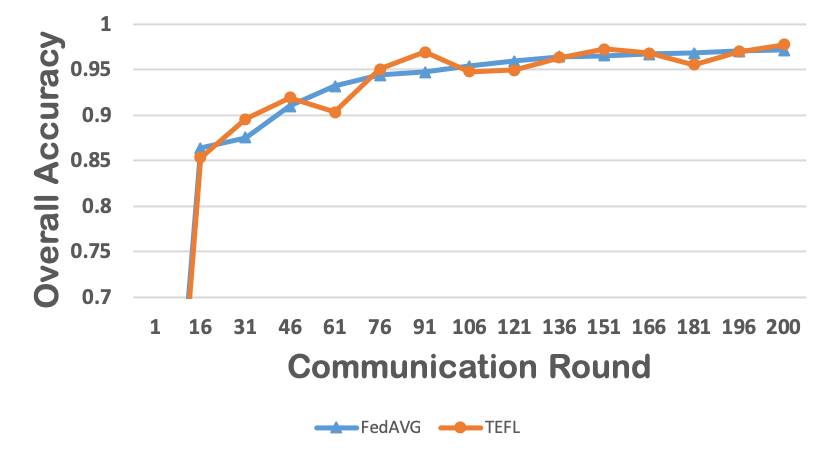}}
		\caption{Comparative studies on temporally weighted aggregation on dataset MNIST using the CNN.}
		\label{fig_9}
	\end{figure}
	
	\begin{figure}[h]
		\centering
		\subfigure[1@HAR.]{
			\includegraphics[height=1.86in]{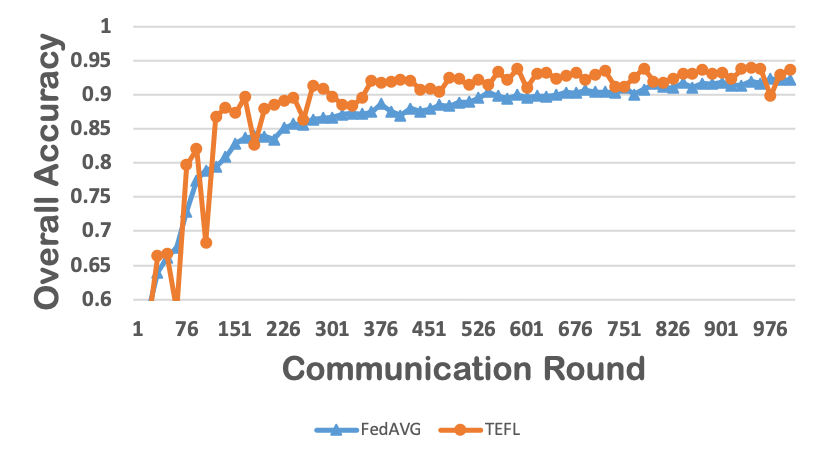}}
		\subfigure[2@HAR.]{
			\includegraphics[height=0.86in]{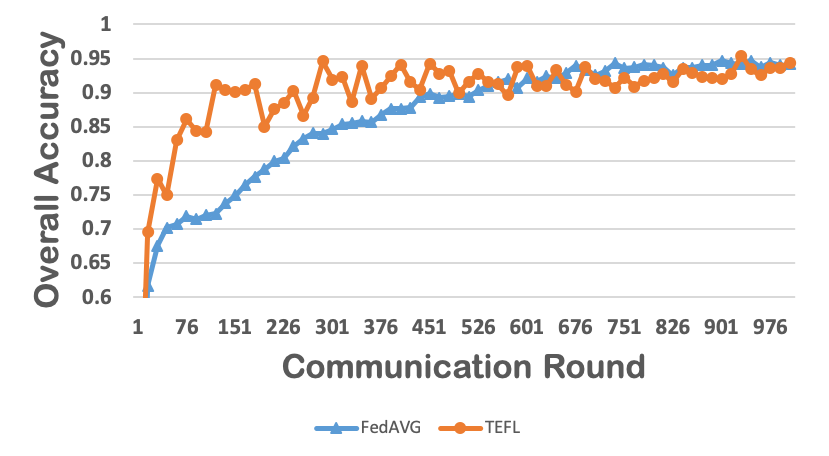}}
		\subfigure[3@HAR.]{
			\includegraphics[height=0.86in]{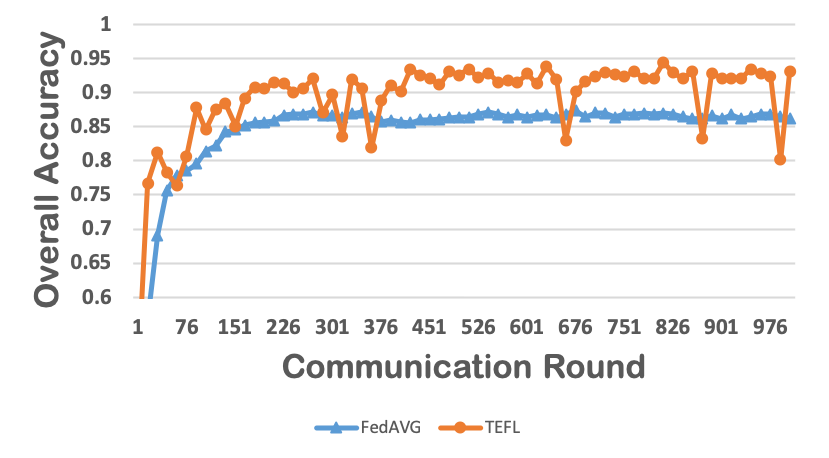}}
		\subfigure[4@HAR.]{
			\includegraphics[height=0.86in]{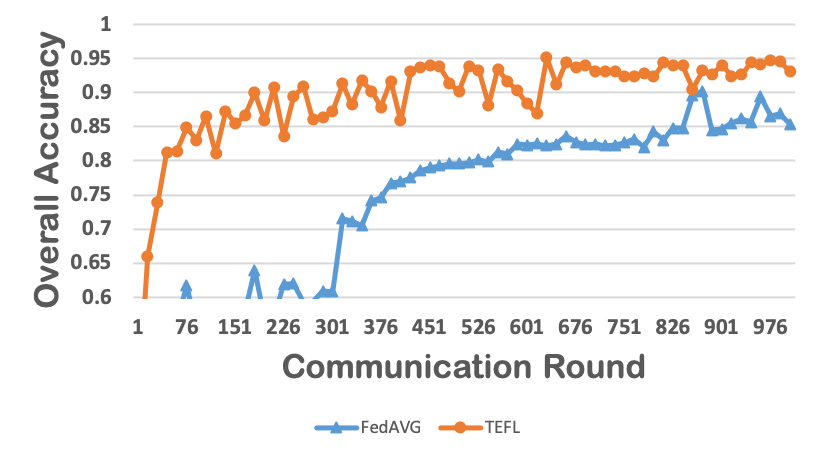}}
		\subfigure[5@HAR.]{
			\includegraphics[height=0.86in]{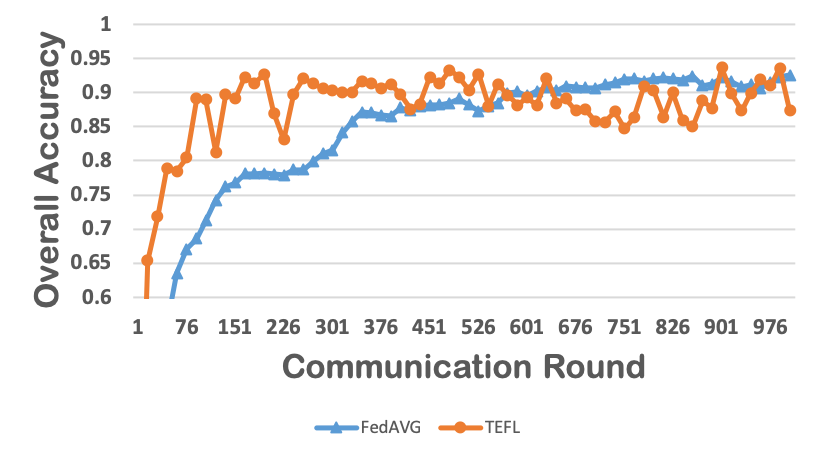}}
		\caption{Comparative studies on temporally weighted aggregation on dataset HAR using the LSTM.}
		\label{fig_10}
	\end{figure}
	
	The following experiments are conducted for testing the overall performance of the algorithms under comparison using the default values of the parameters. Five local datasets are predefined for MNIST and HAR, respectively. For instance, local dataset 1 of task MNIST are termed as 1@MNIST. The importance of the temporal weighting is first demonstrated by comparing the changes of the accuracy over the computing rounds. The baseline FedAVG and TEFL without the asynchronous update are compared and the results on the MNIST and HAR datasets are shown in Fig.\ref{fig_9} and Fig. \ref{fig_10}, respectively. 
	
	The following conclusions can be reached from the results in the two figures. First, the proposed temporally weighted aggregation helps the central model to converge to an acceptable accuracy. On the 1@MNIST and 2@MNIST datasets, TEFL needs about 30 communication rounds to achieve an accuracy reaches to 95.0\%, while the traditional FedAVG needs about 75 rounds to achieve a similar accuracy, leading to a reduce of 40\% communication cost. Similar conclusions can also be drawn on datasets 3@MNIST, 4@MNIST and 5@MNIST, although the accuracy of TEFL becomes more fluctuating. Second, on the HAR dataset, a more difficult task, TEFL can mostly achieve a higher accuracy than FedAVG except on 5@HAR. Notably, TEFL only needs about 75 communication rounds to achieve an accuracy of 90\%, while FedAVG requires about 750 rounds, resulting in a nearly 90\% reduction of the communication cost. Even on 5@HAR, TEFL shows a much faster convergence than FedAVG in the early stage. Finally, the temporally weighted aggregation may result in some fluctuations on the learning performance, which may be attributed to the fact that the contributions of some high-quality local models are less weighted in some communication rounds. \par
	
	\begin{table*}[!t]
		\centering
		\caption{Experiments on Performance .}
		\label{Table10}
		\begin{threeparttable}
			\begin{tabular}{@{}ccccc@{}}
				\toprule
				\textbf{Dataset\_ID@Task} &
				\begin{tabular}{@{}c@{}}
					\textbf{FedAVG} \\
					\midrule
					\begin{tabular}{@{}cc@{}}
						\textit{\textbf{Round (Accuracy)}\tnote{$\star$}} & \textit{\textbf{C. Cost}\tnote{$\star\star$}} \\
					\end{tabular}
				\end{tabular} & 
				\begin{tabular}{@{}c@{}}
					\textbf{TEFL} \\
					\midrule
					\begin{tabular}{@{}cc@{}}
						\textit{\textbf{Round (Accuracy)}\tnote{$\star$}} & \textit{\textbf{C. Cost}\tnote{$\star\star$}} \\
					\end{tabular}
				\end{tabular}  & 
				\begin{tabular}{@{}c@{}}
					\textbf{TWAFL} \\
					\midrule
					\begin{tabular}{@{}cc@{}}
						\textit{\textbf{Round (Accuracy)}\tnote{$\star$}} & \textit{\textbf{C. Cost}\tnote{$\star\star$}} \\
					\end{tabular}
				\end{tabular} &
				\begin{tabular}{@{}c@{}}
					\textbf{AFL} \\
					\midrule
					\begin{tabular}{@{}cc@{}}
						\textit{\textbf{Round (Accuracy)}\tnote{$\star$}} & \textit{\textbf{C. Cost}\tnote{$\star\star$}} \\
					\end{tabular}
				\end{tabular} \\ 
				\midrule
				1@MNIST\tnote{*} & 
				\begin{tabular}{@{}cc@{}}
					75 (97.2\%) & 6.16 \\
				\end{tabular} & 
				\begin{tabular}{@{}cc@{}}
					\textbf{31 (97.9\%)} & 0.74 \\
				\end{tabular} &
				\begin{tabular}{@{}cc@{}}
					106 (97.7\%) & 1 \\
				\end{tabular} &
				\begin{tabular}{@{}cc@{}}
					175 (95.4\%) & 2.46 \\
				\end{tabular} \\
				2@MNIST\tnote{*} & 
				\begin{tabular}{@{}cc@{}}
					85 (97.2\%) &  6.76 \\
				\end{tabular} & 
				\begin{tabular}{@{}cc@{}}
					\textbf{32 (98.5\%)} &  1.33 \\
				\end{tabular} &
				\begin{tabular}{@{}cc@{}}
					61 (98.1\%) & \textbf{1} \\
				\end{tabular} &
				\begin{tabular}{@{}cc@{}}
					---(94.8\%)$\dagger$  & ---$\dagger$ \\
				\end{tabular} \\
				3@MNIST\tnote{*} & 
				\begin{tabular}{@{}cc@{}}
					73 (97.7\%) &  5.99 \\
				\end{tabular} & 
				\begin{tabular}{@{}cc@{}}
					\textbf{31 (98.7\%)} & 1.13 \\
				\end{tabular} &
				\begin{tabular}{@{}cc@{}}
					70 (97.9\%) & \textbf{1} \\
				\end{tabular} &
				\begin{tabular}{@{}cc@{}}
					---(94.9\%)$\dagger$  & ---$\dagger$ \\
				\end{tabular} \\
				4@MNIST\tnote{*} & 
				\begin{tabular}{@{}cc@{}}
					196 (95.2\%) & 8.18 \\
				\end{tabular} & 
				\begin{tabular}{@{}cc@{}}
					\textbf{61 (98.0\%)} & 1.14  \\
				\end{tabular} &
				\begin{tabular}{@{}cc@{}}
					136 (96.1\%) & \textbf{1} \\
				\end{tabular} &
				\begin{tabular}{@{}cc@{}}
					---(93.1\%)$\dagger$  & ---$\dagger$  \\
				\end{tabular} \\
				5@MNIST\tnote{*} & 
				\begin{tabular}{@{}cc@{}}
					98 (97.1\%) & 3.28  \\
				\end{tabular} & 
				\begin{tabular}{@{}cc@{}}
					\textbf{76 (97.8\%)} & 3.17  \\
				\end{tabular} &
				\begin{tabular}{@{}cc@{}}
					61 (96.1\%) &  \textbf{1} \\
				\end{tabular} &
				\begin{tabular}{@{}cc@{}}
					109 (96.7\%) & 2.66 \\
				\end{tabular} \\
				1@HAR\tnote{**} & 
				\begin{tabular}{@{}cc@{}}
					526 (92.3\%) & 4.72 \\
				\end{tabular} & 
				\begin{tabular}{@{}cc@{}}
					\textbf{166} (94.0\%) & \textbf{0.69} \\
				\end{tabular} &
				\begin{tabular}{@{}cc@{}}
					358 \textbf{(94.6\%)} & 1 \\
				\end{tabular} &
				\begin{tabular}{@{}cc@{}}
					---(89.2\%)  $\ddagger$ & ---$\ddagger$ \\
				\end{tabular} \\
				2@HAR\tnote{**} & 
				\begin{tabular}{@{}cc@{}}
					451 (94.7\%) &  5.65 \\
				\end{tabular} & 
				\begin{tabular}{@{}cc@{}}
					\textbf{119} (95.4\%)  & \textbf{0.57} \\
				\end{tabular} &
				\begin{tabular}{@{}cc@{}}
					313 \textbf{(95.9\%)}  & 1 \\
				\end{tabular} &
				\begin{tabular}{@{}cc@{}}
					---(82.9\%)  $\ddagger$ & ---$\ddagger$ \\
				\end{tabular} \\
				3@HAR\tnote{**} & 
				\begin{tabular}{@{}cc@{}}
					---(87.3\%)  $\ddagger$ & ---$\ddagger$ \\
				\end{tabular} & 
				\begin{tabular}{@{}cc@{}}
					\textbf{174 (94.4\%)} & \textbf{0.69} \\
				\end{tabular} &
				\begin{tabular}{@{}cc@{}}
					376 (93.2\%) & 1 \\
				\end{tabular} &
				\begin{tabular}{@{}cc@{}}
					---(88.5\%) $\ddagger$ & ---$\ddagger$ \\
				\end{tabular} \\
				4@HAR\tnote{**} & 
				\begin{tabular}{@{}cc@{}}
					856 (90.2\%)  & 7.05  \\
				\end{tabular} & 
				\begin{tabular}{@{}cc@{}}
					\textbf{181 (95.1\%)} & 1.28  \\
				\end{tabular} &
				\begin{tabular}{@{}cc@{}}
					211 (94.1\%)  & \textbf{1} \\
				\end{tabular} &
				\begin{tabular}{@{}cc@{}}
					751 (91.2\%)  & 5.30 \\
				\end{tabular} \\
				5@HAR\tnote{**} & 
				\begin{tabular}{@{}cc@{}}
					571 (92.6\%)  & 5.49  \\
				\end{tabular} & 
				\begin{tabular}{@{}cc@{}}
					\textbf{155 (93.6\%)} & \textbf{0.57}  \\
				\end{tabular} &
				\begin{tabular}{@{}cc@{}}
					404 (93.1\%) & 1 \\
				\end{tabular} &
				\begin{tabular}{@{}cc@{}}
					646 (92.5\%)  & 2.38 \\
				\end{tabular} \\
				\bottomrule
			\end{tabular}
			\begin{tablenotes}
				\item[*] Cells are filled when the accuracy reaches 95\% within 200 rounds.
				\item[**] Cells are filled when the accuracy reaches 90\% within 1000 rounds.
				\item[$\dagger$] Cells are marked when it can not reach 95\% within 200 rounds.
				\item[$\ddagger$] Cells are marked when it can not reach 90\% within 1000 rounds.
				\item[$\star$] Rounds are needed to reach a certain accuracy, after which the best accuracy (within 200/1000 rounds) is listed in parenthesis.
				\item[$\star\star$] Total communication cost; the C. cost of TWAFL is termed as 1.
			\end{tablenotes}
		\end{threeparttable}
	\end{table*}
	
	Finally, the comparative results of the four algorithms on ten test cases generated from MNIST and HAR tasks are given in Table \ref{Table10}. The listed metrics include the number of rounds needed, the classification accuracy (listed in parenthesis), and total communication cost (C. Cost for short). From these results, the following observations can be made: 
	\begin{itemize}
		\item Both TWAFL and TEFL outperform FedAVG on most cases in terms of the total number of rounds, the best accuracy, and the total communication cost. \par    
		\item TEFL achieves the best performance on most tasks in terms of the total number of rounds and the best accuracy. The temporally weighted aggregation strategy accelerates the convergence of the learning and improved the learning performance.  \par 
		\item TWAFL performs slightly better than TEFL on MNIST in terms of the total communication cost, while TEFL works better than TWAFL on the HAR datasets. The asynchronous model update strategy significantly contributes to reducing the communication cost per round. \par   
		\item AFL performs the worst among the four compared algorithms. When comparing the performance of the AFL only adopting the asynchronous strategy and the TWAFL using both of them, the asynchronous one always needs the help of the other strategy. \par    
	\end{itemize}	
	
	\section{CONCLUSIONS AND FUTURE WORK}\label{sec5}
	This paper aims to reduce the communication costs and improve the learning performance of federated learning by suggesting an synchronous model update strategy and a temporally weighted aggregation method. Empirical studies comparing the performance and communication costs of the canonical federated learning and the proposed federated learning on the MNIST and human action recognition datasets demonstrate that the proposed asynchronous federated learning with temporally weighted aggregation outperforms the canonical one in terms of both learning performance and communication costs.
	
	This study follows the assumption that all local models adopt the same neural networks architecture and share same hyper parameters such as the learning rate of SGD. In future research, we are going to develop new federated learning algorithms allowing clients to evolve their local models to further improve the learning performance and reduce the communication costs.
	
	\section*{Acknowledgment}
	This work is supported by the National Natural Science Foundation of China with Grant No.61473298 and 61876184.
	
	\ifCLASSOPTIONcaptionsoff
	\newpage
	\fi
	
	\bibliographystyle{IEEEtran}

\begin{thebibliography}{10}
\providecommand{\url}[1]{#1}
\csname url@samestyle\endcsname
\providecommand{\newblock}{\relax}
\providecommand{\bibinfo}[2]{#2}
\providecommand{\BIBentrySTDinterwordspacing}{\spaceskip=0pt\relax}
\providecommand{\BIBentryALTinterwordstretchfactor}{4}
\providecommand{\BIBentryALTinterwordspacing}{\spaceskip=\fontdimen2\font plus
\BIBentryALTinterwordstretchfactor\fontdimen3\font minus
  \fontdimen4\font\relax}
\providecommand{\BIBforeignlanguage}[2]{{%
\expandafter\ifx\csname l@#1\endcsname\relax
\typeout{** WARNING: IEEEtran.bst: No hyphenation pattern has been}%
\typeout{** loaded for the language `#1'. Using the pattern for}%
\typeout{** the default language instead.}%
\else
\language=\csname l@#1\endcsname
\fi
#2}}
\providecommand{\BIBdecl}{\relax}
\BIBdecl

\bibitem{mcmahan2017communication}
B.~McMahan, E.~Moore, D.~Ramage, S.~Hampson, and B.~A. y~Arcas,
  ``Communication-efficient learning of deep networks from decentralized
  data,'' in \emph{Artificial Intelligence and Statistics}, 2017, pp.
  1273--1282.

\bibitem{Konecny2015}
J.~Kone{\v{c}}n{\'{y}}, B.~McMahan, and D.~Ramage, ``{Federated Optimization:
  Distributed Optimization Beyond the Datacenter},'' \emph{arXiv Prepr.
  arXiv1511.03575}, no.~1, pp. 1--5, 2015.

\bibitem{Konecny2016}
J.~Konecn{\'{y}}, H.~B. McMahan, F.~X. Yu, P.~Richt{\'{a}}rik, A.~T. Suresh,
  and D.~Bacon, ``{Federated Learning: Strategies for Improving Communication
  Efficiency},'' \emph{CoRR}, vol. abs/1610.0, no. Nips, pp. 1--5, 2016.

\bibitem{lecun2015deep}
Y.~LeCun, Y.~Bengio, and G.~Hinton, ``Deep learning,'' \emph{nature}, vol. 521,
  no. 7553, p. 436, 2015.

\bibitem{Shin2016Deep}
H.~C. Shin, H.~R. Roth, M.~Gao, L.~Lu, Z.~Xu, I.~Nogues, J.~Yao, D.~Mollura,
  and R.~M. Summers, ``Deep convolutional neural networks for computer-aided
  detection: Cnn architectures, dataset characteristics and transfer
  learning,'' \emph{IEEE Transactions on Medical Imaging}, vol.~35, no.~5, p.
  1285, 2016.

\bibitem{Greff2015LSTM}
K.~Greff, R.~K. Srivastava, J.~Koutník, B.~R. Steunebrink, and J.~Schmidhuber,
  ``Lstm: A search space odyssey,'' \emph{IEEE Transactions on Neural Networks
  and Learning Systems}, vol.~28, no.~10, pp. 2222--2232, 2015.

\bibitem{ma2017distributed}
C.~Ma, J.~Kone{\v{c}}n{\`y}, M.~Jaggi, V.~Smith, M.~I. Jordan,
  P.~Richt{\'a}rik, and M.~Tak{\'a}{\v{c}}, ``Distributed optimization with
  arbitrary local solvers,'' \emph{Optimization Methods and Software}, vol.~32,
  no.~4, pp. 813--848, 2017.

\bibitem{reddi2016aide}
S.~J. Reddi, J.~Kone{\v{c}}n{\`y}, P.~Richt{\'a}rik, B.~P{\'o}cz{\'o}s, and
  A.~Smola, ``Aide: fast and communication efficient distributed
  optimization,'' \emph{arXiv preprint arXiv:1608.06879}, 2016.

\bibitem{shamir2014communication}
O.~Shamir, N.~Srebro, and T.~Zhang, ``Communication-efficient distributed
  optimization using an approximate newton-type method,'' in
  \emph{International conference on machine learning}, 2014, pp. 1000--1008.

\bibitem{zhang2015disco}
Y.~Zhang and X.~Lin, ``Disco: Distributed optimization for self-concordant
  empirical loss,'' in \emph{International conference on machine learning},
  2015, pp. 362--370.

\bibitem{chilimbi2014project}
T.~M. Chilimbi, Y.~Suzue, J.~Apacible, and K.~Kalyanaraman, ``Project adam:
  Building an efficient and scalable deep learning training system.'' in
  \emph{OSDI}, vol.~14, 2014, pp. 571--582.

\bibitem{dean2012large}
J.~Dean, G.~Corrado, R.~Monga, K.~Chen, M.~Devin, M.~Mao, A.~Senior, P.~Tucker,
  K.~Yang, Q.~V. Le \emph{et~al.}, ``Large scale distributed deep networks,''
  in \emph{Advances in neural information processing systems}, 2012, pp.
  1223--1231.

\bibitem{konevcny2016federated}
J.~Kone{\v{c}}n{\`y}, H.~B. McMahan, D.~Ramage, and P.~Richt{\'a}rik,
  ``Federated optimization: Distributed machine learning for on-device
  intelligence,'' \emph{arXiv preprint arXiv:1610.02527}, 2016.

\bibitem{Boyd2011a}
S.~Boyd, N.~Parikh, E.~Chu, B.~Peleato, and J.~Eckstein, ``{Distributed
  Optimization and Statistical Learning via the Alternating Direction Method of
  Multipliers},'' \emph{Foundations and Trends in Machine Learning}, vol.~3,
  no.~1, pp. 1--122, 2011.

\bibitem{yosinski2014transferable}
J.~Yosinski, J.~Clune, Y.~Bengio, and H.~Lipson, ``{How transferable are
  features in deep neural networks?}'' in \emph{Advances in neural information
  processing systems}, 2014, pp. 3320--3328.

\bibitem{geyer2017differentially}
R.~C. Geyer, T.~Klein, and M.~Nabi, ``Differentially private federated
  learning: A client level perspective,'' \emph{arXiv preprint
  arXiv:1712.07557}, 2017.

\bibitem{bonawitz2017practical}
K.~Bonawitz, V.~Ivanov, B.~Kreuter, A.~Marcedone, H.~B. McMahan, S.~Patel,
  D.~Ramage, A.~Segal, and K.~Seth, ``Practical secure aggregation for
  privacy-preserving machine learning,'' in \emph{Proceedings of the 2017 ACM
  SIGSAC Conference on Computer and Communications Security}.\hskip 1em plus
  0.5em minus 0.4em\relax ACM, 2017, pp. 1175--1191.

\bibitem{Zhu2018}
H.~Zhu and Y.~Jin, ``Multi-objective evolutionary federated learning,'' 2018,
  •.

\bibitem{krizhevsky2014one}
A.~Krizhevsky, ``{One weird trick for parallelizing convolutional neural
  networks},'' \emph{arXiv preprint arXiv:1404.5997}, 2014.

\bibitem{krizhevsky2012imagenet}
A.~Krizhevsky, I.~Sutskever, and G.~E. Hinton, ``Imagenet classification with
  deep convolutional neural networks,'' in \emph{Advances in neural information
  processing systems}, 2012, pp. 1097--1105.

\bibitem{lecun1998gradient}
Y.~LeCun, L.~Bottou, Y.~Bengio, and P.~Haffner, ``Gradient-based learning
  applied to document recognition,'' \emph{Proceedings of the IEEE}, vol.~86,
  no.~11, pp. 2278--2324, 1998.

\bibitem{anguita2013public}
D.~Anguita, A.~Ghio, L.~Oneto, X.~Parra, and J.~L. Reyes-Ortiz, ``A public
  domain dataset for human activity recognition using smartphones.'' in
  \emph{ESANN}, 2013.

\bibitem{gers1999learning}
F.~A. Gers, J.~Schmidhuber, and F.~Cummins, ``Learning to forget: Continual
  prediction with lstm,'' 1999.

\end{thebibliography}

\end{document}